# Emergent AI-assisted discourse: Case study of a second language writer authoring with ChatGPT

Sharin R. Jacob,* Tamara Tata, Mark Warschauer

*University of California, Irvine, Irvine, United States*

*Corresponding author: sharinj@uci.edu

# Emergent AI-assisted discourse: Case study of a second language writer authoring with ChatGPT

The rapid proliferation of ChatGPT has incited debates regarding its impact on human writing. Amid concerns about declining writing standards, this study investigates the role of ChatGPT in facilitating academic writing, especially among language learners. Using a case study approach, this study examines the experiences of Kailing, a doctoral student, who integrates ChatGPT throughout their academic writing process. The study employs activity theory as a lens for understanding writing with generative AI tools and data analyzed includes semi-structured interviews, writing samples, and GPT logs. Results indicate that Kailing effectively collaborates with ChatGPT across various writing stages while preserving her distinct authorial voice and agency. This underscores the potential of AI tools such as ChatGPT to enhance academic writing for language learners without overshadowing individual authenticity. This case study offers a critical exploration of how ChatGPT is utilized in the academic writing process and the preservation of a student's authentic voice when engaging with the tool.

**Introduction**

The release and rapid diffusion of ChatGPT has sparked wildly different views of its relationship to human writing.  In light of longstanding concerns about a national writing crisis (Intersegmental Committee, 2002; The National Commission on Writing, 2003, 2004), educators have voiced apprehensions that the teaching of writing and its use as an assessment tool might be seriously compromised. Conversely, the business sector eagerly anticipated the potential for rapidly producing written content at a reduced cost.

Historically, students in the US have grappled with mastering academic writing genres (Graham & Perin, 2007). Data from the National Assessment of Educational Progress (NAEP) revealed that over 20% of 12th-grade students were *below basic* writers. Disaggregated data showed that only 13% of White students were *below basic*, while 35% of Hispanic and 39% of Black students were *below basic* (National Center for Education Statistics, 2012). English learners face heightened challenges with their writing as they reflect a wealth of cultural and social values that are too often undervalued by mainstream academic settings (Booth et al., 2023; Bunch, 2013).

There is a growing body of work on how second language learners navigate the types of writing necessary for academic research (Curtis, 2023; Rahaman et al., 2023; Steiss et al., 2023; Tseng & Warschauer, M., 2023; Warschauer et al., 2023). The value of ChatGPT, in offering input, feedback, or scaffolding for writing, is increasingly acknowledged. Yet, a pivotal question arises: Does relying on AI-generated texts equate to academic misconduct, or can AI-assisted writing coexist with the preservation of the human author's distinct voice and agency?

To explore this issue, we present a case study of a student who thoroughly integrates ChatGPT into all aspects of the authoring process. First, we provide a background on using ChatGPT to facilitate writing for language learners. Next, we describe our participant, Kailing, who uses ChatGPT throughout each stage of the writing process, to write an academic proposal during her doctoral studies. We then describe how activity theory can be employed as a theoretical framework to advance our understanding of how AI tools such as ChatGPT can mediate the writing process. From there, we describe our single-subject case study design and data collection, including semi-structured interviews, analysis of writing artifacts, and GPT logs. Findings from our case study indicate that Kailing applied her expertise to dynamically interact with ChatGPT during the brainstorming, research, text generation, and revision processes. Notably, throughout this process, she retained her own sense of voice and agency. Findings such as these highlight the potential of generative AI tools, such as ChatGPT, to enhance writing for language learners in academic settings.

This paper addresses the following research questions:

1) How does a graduate student and second language writer in English employ ChatGPT in each phase of her academic writing process?

2) To what extent does the student maintain her authentic voice when utilizing ChatGPT in the writing process, and how is her voice characterized within this context?

**Background**

In the past two years, there has been a rapid increase in studies on the integration of AI in language learning (Bin-Hady et al., 2023; Fitria et al., 2023; Kohnke et al., 2023; Liang et al., 2023; Steiss et al., 2023; Tseng et al., 2023; Warschauer et al., 2023). Although most findings indicate that language learners could greatly benefit from generative AI tools, they face disproportionate challenges regarding accusations of academic misconduct. Notably, GPT detectors exhibit considerable bias against these students. A study that analyzed the precision of seven widely used GPT detectors on 91 TOEFL essays and 88 US student essays found that while the detectors reliably identified essays from US students, they mistakenly labeled over half of the TOEFL essays as "AI generated," with an average false-positive rate of 61.3% (Liang et al., 2023). Building upon this work, Warschauer et al. (2023) explore the affordances and challenges of using AI-based tools for second language learners. They pinpoint the dilemmas faced by these learners as they engage with AI tools. To echo Liang et al. (2023), they underscore the pressure on language learners to mirror the target language. However, when L2 students utilize tools such as ChatGPT, they often face allegations of plagiarism, especially when compared to native speakers (Liang et al., 2023).

In addition to detection bias, generative AI tools, such as ChatGPT, have been shown to exhibit discriminatory bias (Kenthapadi et al., 2023). Since they are trained on large corpora of data, engineers encounter challenges when attempting to audit the data for specific biases and discriminatory elements (Bender et al., 2021). For instance, large language models have exhibited biases and stereotypes related to gender (Bolukbasi et al., 2016), disability (Hutchinson et al., 2020), and language (Lee, 2023). Given these findings, it is imperative to approach the use and interpretation of outputs from generative AI tools with caution, ensuring that measures are in place to counteract and mitigate inherent biases.

There is substantial evidence indicating that ChatGPT can perpetuate bias, stereotypes, and inaccuracies. However, despite its recognized limitations, it would be a mistake to overlook its potential benefits for underperforming students, such as emerging writers, students with disabilities, and language learners. Regarding job readiness, Warschauer et al. (2023) highlight the eagerness of corporations to incorporate AI into their systems, yet students designated as English learners have limited access and opportunities to learn how to effectively utilize these tools. Denying language learners the chance to learn and employ generative AI tools can adversely impact their job readiness. A recent study found that using ChatGPT for work-related tasks in areas such as marketing and grant writing reduced the time to complete tasks by 40% while improving quality by 20% according to a double-blinded review of work performed. Importantly, low-performing employees improved the most (Noy & Zhang, 2023). A similar study from Harvard Business School assigned strategy consultants to conceptualize and develop new product ideas in one of three conditions–without AI, with ChatGPT, and with ChatGPT and an overview of prompt engineering. Again, the two ChatGPT groups improved the most, largely due to a catch-up effect among the below-average writers who improved by 43% in their writing tasks compared to an improvement of 17% among above-average writers (Dell'Acqua et al., 2023). Findings such as these suggest that learning tools such as ChatGPT have the potential to level the playing field for struggling writers in the workforce.

Many are concerned that generative AI tools such as ChatGPT could boost productivity at the expense of innovation and creativity. However, empirical evidence suggests that when utilized appropriately, that may not be the case. A study investigating the influence of GenAI on creative content production found that individuals inspired by AI-generated content crafted more innovative narratives compared to those drawing from other sources (Doshi & Hauser, 2023). This impact increased significantly among writers perceived as less creative. Such results highlight ChatGPT's potential to equalize opportunities for emerging writers.

There are legitimate concerns that technology cannot replicate the nuances of advanced writing, tasks such as these require creativity, innovation, high-quality content, structured organization, and logically sound arguments. However, a recent study investigating ChatGPT's role in supporting students with argumentative suggested that ChatGPT aids in content generation and streamlines the writing process, including outline preparation, content revision, proofreading, and post-writing reflection (Su et al., 2023). Leveraging generative AI tools for complex writing tasks, such as argumentative writing, is particularly beneficial for language learners, who often grapple more with the higher-order components of argumentative writing compared to their native-speaking counterparts (Booth et al., 2023).

Finally, it has been long argued that human feedback is irreplaceable. However, emerging studies show generative AI tools such as ChatGPT are capable of delivering high-quality feedback for students. For example, a current investigation comparing feedback from humans to that provided by ChatGPT revealed that the practical differences between the two were minimal (Steiss et al., 2023). The consistency between human and AI-generated feedback suggests that ChatGPT could be a reliable and scalable tool, potentially bridging the feedback gap often faced by language learners due to limited human resources.

In conclusion, the rapid integration of AI into language learning has opened new horizons, offering potential benefits, especially for underperforming and L2 students. However, the challenges—ranging from detection and discriminatory biases to issues of academic integrity and the potential dilution of creativity—are undeniable. These concerns underscore the importance of a balanced and cautious approach to the application of tools such as ChatGPT. Notwithstanding its limitations, the evidence suggests that, when used judiciously, ChatGPT can be an asset, providing quality feedback, fostering creativity, and aiding in content generation. Moving forward, as the field continues to evolve, continuous evaluation and mindful implementation of such tools will be crucial to ensure that they serve to augment rather than impede the processes of language learning and advanced writing.

We now turn to the participant in our case study to better understand the conditions under which she employed ChatGPT to produce academic research.

**The writer**

Kailing (pseudonym) is an international graduate student from China at a research university in the United States. She has been in the US for four years for her master's studies and the beginning part of her doctoral study. Kailing is moderately proficient in English, especially considering that she didn't major in English as an undergraduate, but like most international students, she is not yet a polished English language writer. Kailing is a skilled and comfortable user of new technologies and has long used other tools, such as Grammarly, to help her with her writing.

Kailing came to our attention as an extraordinarily capable writer for a learner of English. She received straight As or A+s in her first year of doctoral studies in a writing-intensive field, won a prestigious best paper award at a highly competitive international conference, and has authored grant proposals deemed by faculty to be at a quality level as high or higher than those written by a typical research professor.

It is not only Kailing's writing that brought her to our attention but also her comprehensive use of ChatGPT at all stages of the writing process. After a brief interview with Kailing, it became clear that she was the most thorough and proficient user of ChatGPT for academic writing that we had met or knew of. We then invited her to be part of this case study by answering our interview questions and sharing with us her prompts and ChatGPT responses as well as the iterative versions of her papers– and generally telling and showing us how she uses the tool for academic writing. We believe that it is important for academics debating the use of generative AI to have a transparent look at how one highly effective student uses the technology before deciding to ban AI in their classes; rather this case study might inspire some use cases to explore for other students.

**Theoretical framework**

Activity theory, anchored in Engeström's (1987, 1999) work and drawing upon Vygotsky's (1978) sociocultural framework, posits that human interactions are multifaceted, influenced by numerous factors, and inherently dynamic. In aligning this theory with the writing process, we lean on Kessler's (2020) conceptualization. Within this framework: The *subject* (or agent) undertakes actions with a specific goal in mind. These actions are facilitated by tools or *mediators*. In the realm of writing, the end product (the written text) is the goal, while various aids and resources (e.g., computer, chatbot, editing software, thesaurus) facilitate the creation of this text. The interaction is also shaped by three contextual elements: *Rules* provide guidelines or standards for the subject's behavior. *Community* represents the broader context and stakeholders that influence the subject's actions. *Division of labor* illustrates how tasks related to the primary goal are distributed among participants. In the context of academic writing, a researcher (subject) may engage with colleagues, field experts, editors, and peers during the writing and revision process. Notably, tensions might arise, such as discrepancies between established writing norms and specific assignment criteria. These tensions, however, can serve as catalysts for the learner's developmental journey (Fujioka, 2014).

We build on this theory by contextualizing it within the use of generative AI tools for writing. In our case study, the subject undertakes the goal of writing a research proposal. The end product, or written text, is mediated by ChatGPT, a generative AI tool that uses language processing to facilitate writing, among other tasks. Within this dynamic interaction between text and tool, our participant is constrained by the rules of academic writing, and the context of her broader community (i.e., the reviewers of the proposal), while engaging in the division of labor by conferring with her colleagues, peers, and scholars on how to improve her writing.

Our study is also based on the cognitive process models of writing (Flower & Hayes, 1981), in which writing consists of planning, translating, and reviewing, and revising. Composition is a recursive process (McCutchen, 1996; Berninger et al., 1996): writers cycle through the planning, translating, and reviewing multiple times, and these stages all interact with one another throughout the composing process (Flower & Hayes, 1981).

**Materials and methods**

*Research design*

A case study (Mabry, 2008) design was employed to delve deeply into the specific experiences of a single participant. This approach aimed to elucidate the ways in which the participant utilizes ChatGPT for academic research, emphasizing her retention of voice and agency.

*Participant*

The study centered on Kailing; a PhD student enrolled in a graduate program in Southern California. She was intentionally selected for her unique engagement with ChatGPT in the context of academic research.

*Data instruments*

To explore the strategies employed by Kailing in her academic research with ChatGPT, we utilized semi-structured interviews. These interviews allowed for both predetermined questions and flexibility for follow-up queries based on Kailing's responses. A total of four interviews, totaling about two hours, were conducted. The first three interviews were conducted over the summer to understand how Kailing used ChatGPT for the purpose of academic writing. We then conducted a second interview during late fall to better understand how her use of ChatGPT changed over time. We also analyzed her written artifacts, including her research papers, proposals, and ChatGPT logs.

*Data analysis*

The analysis of interview data was undertaken employing inductive qualitative coding approaches (Saldaña, 2021). The first cycle of coding involved identifying themes that pertained to the research questions and were rooted in literature to date on the use of ChatGPT for language learners. After assigning initial codes, we developed a codebook to reveal emergent themes from the study. To enhance the validity, we conducted member checks by sharing our interpretations with the participant for their feedback (Lincoln & Guba, 1985). This case study focused on a singular participant to explore the nuanced interactions between generative AI and language acquisition processes.

**Results**

*The writing*

Some people use ChatGPT to brainstorm ideas or outlines for their writing. Others use it as a very capable copy editor. In either case, they claim their writing is authentic, rather than plagiarized, because ChatGPT is only used for 1-2 steps, and not for drafting the content. In contrast, Kailing uses ChatGPT for brainstorming and outlining, for editing, for *drafting*–and for much else in between. Yet she harnesses such control of every step of the process that we consider what she is doing as authoring, even if it is not writing in the traditional sense. The best way to understand that is by example, so below we present a step-by-step analysis of how Kailing wrote one of her recent outstanding papers, a research proposal.

Kailing's writing process, assisted by ChatGPT, can be categorized into four distinct stages: brainstorming (planning), research text production (translating from thought to text), and revision. Within each stage, Kailing engages in a series of intricate decisions, employing advanced cognitive skills to adapt the text in service of her intended communication goals.

*Brainstorming*

During the brainstorming phase, Kailing progressively refined her queries, moving from broad to specific, based on her existing content area knowledge, knowledge of prompting the AI, and the goals and context of her writing. For instance, while soliciting the chatbot to produce ten instances of how a particular digital technology can be adapted for students with disabilities, she realized that her initial queries were too generic and not feasible for her research context.

> *I kind of realized that if you give a prompt like this to ChatGPT, the ideas I received were too generic and not very applicable to or not very doable to make it into a proposal because it was too broad and if you think deep about the technical details of these ideas, you realize that it is not very feasible.*

To produce more refined ideas, Kailing consulted both her colleagues and existing research literature. In consulting with her peers, Kailing engaged in peer-peer-computer interaction, which benefits language learners by establishing a frame of reference to refine and negotiate meaning in real-time (Bailey & Heritage, 2014; Grapin, 2020; Ruiz-Primo, 2011). Informed by these insights, she adjusted her research query, leading to new discoveries.

> *And then I refined my prompt [sic] ChatGPT and asking, how can I apply conversational agent into the storytelling design with children with ASD? And then ChatGPT told me something that I didn't know….there's this technology called augmentative and alternative communication keyboard for children with autism, which is frequently used for them to help them communicate with people….So I thought was I didn't know this before, but ChatGPT told me this, which I think is a very useful information.*

Drawing on her initial findings, Kailing reviewed scholarly literature to understand the integration of assistive technologies with digital storytelling and conversational agents. This literature review subsequently informed the formulation of her research proposal's core ideas.

> *It's like an interactive storytelling experience with AI where children can use the AC keyboard to …type the starting of the story, and then the system can build upon the children's story based on the graphs that they use to continue this storytelling process. So this is the process of me using ChatGPT to generate ideas for my proposal.*

During the brainstorming phase, Kailing actively refined her prompts through discursive interaction with the chatbot. This discursive interaction involving multiple human participants in conjunction with an AI chatbot empowered Kailing to utilize her own cognitive and social resources to delineate her research (Grapin et al., 2022). However, generative AI has the potential to transcend mere tools for thought, evolving into active participants in the interaction (Grapin et al., 2022; Pierson et al., 2020). From a human interaction perspective, Kailing applied advanced cognitive skills, such as evaluation and analysis, to synthesize her findings and eventually formulate the primary objective of her research proposal. Conversely, the chatbot performed its functions semi-autonomously while operating in tandem with Kailing, its human guide (Brady et al., 2015). While Kailing primarily drove the prompting process with her intellectual input, the task of text generation was delegated to the AI, capitalizing on its superior English language processing capabilities. As "computer programs essentially automate abstractions," (Jacob & Warschauer, 2018, p. 5), GhatGPT effectively amplified the intellectual capacities of Kailing and her colleagues in order to produce a coherent research topic.

*Research*

Kailing employed ChatGPT as a tool for conducting preliminary research and consulting the literature to provide evidence for specific claims. She prompted it for lists of prominent researchers on her topic of interest, sought overarching themes pertinent to her research topic, inquired about researchers particularly aligned with those themes, and further investigated funded projects within her target region that pertained to related subjects.

For example, upon developing an outline for her proposal, she asked ChatGPT to generate citations to support her claims. Understanding the intricacies of how the chatbot functions, Kailing was well-aware that ChatGPT does not always produce credible citations, a recognized limitation of the software.

> *Lots of its arguments cannot be backed up by evidence….This goes to a drawback of ChatGPT which is even though you ask it to include, for example, five or 10 citations to support the literature, it will make up the citations which means it uses fake citations because of how ChatGPT works, which is a probabilistic prediction of the next word based on the previous word. So I think here comes the part where I need to pause and read more literature to…check [what] ChatGPT comes up with if it is accurate or if there is evidence to support it.*

The importance of finding credible citations is underscored by research on citation analysis, which reveals rhetorical markers, discourse structures, epistemological foundations, and empirical perspectives that authors assume when providing a discursive framework of the literature (Hu & Wang, 2014). As researchers agree that incorporating citations into academic research is an especially complex phenomenon (Barks, 2001; Hu & Wang, 2014; Pecorari, 2001), language learners can face challenges when they are asked to integrate textual borrowing strategies into their academic writing (Pecorari, 2001). Furthermore, the consequences for inaccurately citing borrowed materials frequently take on administrative and legal dimensions, leading to significant repercussions for language learners. Recognizing this, Kailing cross-references the citations provided by ChatGPT with the existing literature to ensure their accuracy. By unpacking the complexities of generating citations, Kailing devised a methodology for using ChatGPT to reference the works of others while maintaining academic integrity and sophistication in textual borrowing strategies.

*Text production*

In drafting the text, Kailing adopts an iterative approach with the chatbot. She breaks up her writing sessions into manageable chunks and thinks deeply in advance of each one. For the literature review, for example, she considers the points she wants to make, how to organize them coherently, and comes up with bullet points that she inputs to ChatGPT. ChatGPT then does what for Kailing is the "hard work" as a non-native speaker, it rapidly drafts text covering her points. She then pauses and reads more literature, gathering more content knowledge to inform her writing, and supplements the output. Often, Kailing gives ChatGPT such detailed outlines that it does not seem like the AI is adding much value -- mostly with transitions between ideas, and maybe also confidence and self-efficacy. She has found that these detailed prompts are more useful than simply a broad query based on the assignment. As an illustration, Kailing reported on the following broad query she made regarding the composition of a research proposal.

> *Could you write me the research proposal, including the following sections, which are the sections required by the professor? In this assignment, for example, its introduction, which is like background, and conceptual framework and research question and also methodology. I didn't expect that much from Chat GPT because every time I gave this kind of prompt to ChatGPT, the response was always first of all, very short. And second, especially for the part of the literature review or the conceptual framework. I don't think ChatGPT could tell a very coherent story.*

Kailing recognized that ChatGPT could, at best, supply basic ideas for her proposal. To this end, she drew upon her own expertise to refine and expand on her preliminary draft in order to ensure a more cohesive narrative. In particular, notice her rhetorical awareness of the need to tell a coherent story in order to have a successful proposal. This genre knowledge allows her to successfully note the limitations of ChatGPT's output and also provide more specific prompting to increase the usefulness of the output.

Interestingly, Kailing believes that ChatGPT excels more as a text generator than as a source of high-quality ideas; however, she acknowledges that she continues to glean new insights from the chatbot. In addition, the underlying large language models are evolving rapidly and improving in significant ways, so this may only be a temporary reality.

> *To me, because I think ChatGPT is a very high-quality text generator, but it's…not a very good idea generator. So I always rely on ChatGPT to generate high-quality text instead of using it to generate ideas for me, but on the other hand, I can sometimes get some useful input from chatting material like the AAC keyboard, I wouldn't have known it if there wasn't ChatGPT for me.*

Kailing used ChatGPT for ideation even though it was not adept at producing unique, creative ideas. Nonetheless, even the fallible tool can provide something to react to or refine, and in some cases even provides usable ideas. She then uses the chatbot for its (currently) more skillful ability to produce fluent text, a labor-intensive endeavor, especially for multilingual writers.

> *I think ChatGPT always gives me responses that I'm very satisfied with because it writes something decent and good in a short amount of time, which saves me time.*

Taken together, these findings suggest that ChatGPT acts as an assistive tool, facilitating language learners such as Kailing in expressing ideas in the target language with a speed and efficiency akin to their native language. In this way, natural language processing facilitates human cognitive processing to achieve desired ends, such as producing robust academic writing.

***Revising***

Kailing acknowledges ChatGPT's constraints in delivering the granularity required for scholarly articles. She perceives her initial drafting with ChatGPT as a foundation-building exercise while considering the revision phase as the stage to incorporate essential details and evidentiary support for her assertions.

> *So going back to the prompts that I just talked about, even though I give ChatGPT a seemingly detailed prompt, but still the response that I got from ChatGPT [is] sometimes still too generic and vague. For example, I just asked Chad GPT to write something about the increasing attention paid by the New York Department of Education to early math learning…There is no detail at all….Some really pretty words? Yes. So I would consider this…if I want to write a paragraph about New York's investment in math education, I would consider this as a skeleton of my writing. And then I would for example, go to the New York Department of Education's website to look at its investment and funds to the math program and then add details into the skeleton to make it more [clear].*

During the interview, Kailing repeatedly praises the eloquent language, or the "*really pretty words*" produced by ChatGPT, but also critically assesses the substance of its ideas. Although she concedes having gained some insights from ChatGPT, she believes that the concepts it formulates often fall short of the rigor required for scholarly writing.

*Authorship and voice*

The issue of authorship and voice emerges prominently in Kailing's narrative. While she insists on aligning ChatGPT-generated content with her own voice through revisions, she also values the platform's ability to produce eloquent text—a feature appreciated by both language learners and native speakers. Nevertheless, her distinct authorial identity is shaped by continuous decisions made during her interactions with ChatGPT, parallel discussions with peers, and her engagement with existing academic literature, all aimed at upholding scholarly integrity.

Kailing possesses a distinct understanding of her writing style and voice; consequently, she meticulously revises her work to ensure it authentically mirrors her authentic voice and writing style.

> *But I wouldn't necessarily take what ChatGPT suggests because sometimes I feel **it doesn't look like my writing anymore**. So although I know the language is better, or it used more beautiful words, I don't feel comfortable using something entirely generated by ChatGPT because it wasn't my writing anymore, and so I feel this is against academic ethics, if I use something completely generated by ChatGPT.*

It is interesting that Kailing has such a clear sense of her own voice. Many students have lost their sense of their own voice after high school and the focus on the AP essay genre. Kailing has retained her own voice and guards it fiercely. This is an area that may become more important to teach writers as time goes by and the use of generative AI is commonplace.

Rather than perceiving ChatGPT merely as a writing aid, Kailing regards it as an intellectual collaborator, facilitating her active engagement in the research process.

> *I definitely enjoyed the process more especially because of this thinking partner role of ChatGPT. I feel like I have like a back and forth interaction with someone who's always there. And like, and because of this interaction, I can gradually build upon my previous ideas based on my thinking and ChatGPT input and gradually improve the idea.*

Kailing employed intricate strategies to maintain her authorial integrity while utilizing ChatGPT for academic research. Initially, she engaged with the chatbot as a collaborative tool for refining her ideas, ensuring she did not lean too heavily on it. She meticulously verified the claims and supported them with evidence anchored in extant academic research. Furthermore, Kailing solicited feedback from peers, using her social resources, and immersed herself in existing scholarly works to hone her arguments. Though she incorporated ChatGPT in the drafting stages, she carefully revised its output to achieve the depth and rigor demanded by academic research. Continuous reflection about her writing approach, coupled with a clear recognition of ChatGPT's strengths and weaknesses, allowed her to harness the chatbot's ability to generate high-quality text for her own, specific purposes.

We now turn to Kailing's follow-up interview to examine how her use of ChatGPT for academic writing changed over time.

**Results from follow-up interview**

As Kailing persisted in using ChatGPT, she began to identify predictable patterns in both its content and structure, leading her to decrease her dependence on the program. Specifically, during the brainstorming phase, Kailing observed that regardless of the number of prompts she provided, the software tended to produce the same related topics.

> When I first asked ChatGPT to generate ideas, I got pretty excited about these ideas. But then, after…sitting down and thinking about it for a while, I figured…not all of them are realistic or reliable. Even though I asked [ChatGPT] to elaborate on any of the ideas, I figured that it actually gave me pretty similar things. So I figure that is part of ChatGPT's model. Conversational agents or generative AI is always associated with something like interactive storytelling or adaptive learning. That's how the model works. I think that the words 'interactive storytelling' or' adaptive learning' are more likely to come after 'conversational agents.' So, I started to rely less on it.

In Kailing's experience with ChatGPT, she initially expressed enthusiasm regarding its idea-generation capabilities. However, through her continued use, she began to recognize recurring patterns in the responses. Specifically, she noted that the software frequently presented related suggestions for a given topic. Kailing hypothesized that these patterns are caused by ChatGPT's design, reflecting a predisposition within the model to connect terms such as "conversational agents" with nearby language such as "interactive storytelling" and "adaptive learning." Her recognition of these predictable patterns led her to reduce her reliance on ChatGPT, understanding its limitations in diversifying its content-based suggestions.

Kailing discovered a similar limitation to the AI tool at the sentence level, particularly with word choice.

> [In the past], I would even use some of the sentences [generated by ChatGPT] in my writing. But…since I've used ChatGPT more and more, and also read a lot of text generated by ChatGPT, I've noticed some patterns. For example, [words] like 'harness,' or 'harness AI,' or 'we want to address the significant gap between A and B,' and also when discussing culturally relevant pedagogy. 'We need to pay attention to the cultural nuances.' So some of these signal words are, I noticed, always generated [by ChatGPT]..
> So I started to rely less on it.

Kailing initially incorporated ChatGPT's sentences directly into her work but, over time, recognized repetitive phrasing and vocabulary from the software. These recurring patterns led her to decrease her reliance on the tool for sentence generation.

However, Kailing continued to use ChatGPT to help craft certain textual features at the sentence level that she found to be challenging as a language learner, such as sentence variation.

> I started to rely less on ChatGPT in terms of the framing of the exact sentence or word, but I still would put my writing into ChatGPT and ask it to paraphrase it to help me at more of a sentence level….As a non-native speaker, a challenge I came across during academic writing is it's hard…for me to diversify the sentence structure. It's very easy for me to say, "Oh, past literature demonstrates this. In addition, previous literature also indicates this." So you know, always follow this kind of same sentence structure…makes the writing a little bit boring. But by referring to ChatGPT's paraphrasing, I would know…maybe I can change…the sentence structure a little bit. Like the active tense, the passive tense and how to start the sentence.

Rather than solely relying on ChatGPT for content generation, Kailing began to leverage the tool more specifically for paraphrasing and diversifying her sentences—a challenge she identifies with as a language learner. She acknowledged her tendency to use repetitive sentence structures, which she perceives as monotonous. To this end, ChatGPT represents a valuable tool for language learners, offering targeted support for refining their writing skills, so that these students can better display their content knowledge without being taxed with unnecessary linguistic burdens.

While Kailing still used ChatGPT for multiple purposes in her writing, during the time of the follow-up interview, she found that consulting her colleagues and peers, as well as human writing assistants, was more valuable than using ChatGPT. She explained her reasoning in a joke.

> In software engineering, there is a joke that oh, it looks pretty cool that you can use ChatGPT to generate code. So right now your workflow becomes 1) you get the code from ChatGPT immediately, and then 2) you spend two hours debugging the code generated by chatGPT. I actually feel the same way [with writing].

In the context of academic writing, while tools such as ChatGPT offer immediate support, Kailing's reflection underscores the importance of human expertise. Her analogy between the time-consuming debugging process and the editing and refinement that is necessary to produce high-quality AI-assisted writing emphasizes the irreplaceable value of human expertise in the writing process.

**Discussion**

In this study, we critically engage with two perspectives: the nature and practice of writing in the age of advanced AI, and the implications for academic writers, particularly those situated in contexts comparable to Kailing's.

At the core of human endeavors lies the role of tools. In examining writing through the lens of activity theory, the introduction of a new tool, such as ChatGPT, recalibrates the dynamic of human-tool interaction, necessitating a comprehensive approach that integrates human, tool, and context (Engeström, 1999).

Bateson's (1972) contemplation of the blind man and his stick offers a deeper reflection on this relationship. He posits the inseparability of the tool (the stick) from the user (the blind man), thereby highlighting the symbiotic relationship between them. Similarly, Postman's (1993) observation on the transformation of Europe following the printing press's advent reinforces the transformative potential of tools on societies.

Communication technologies, particularly as highlighted by Postman and Ong (1982), possess the capacity to reconfigure cognitive and societal landscapes. The inception of the printing press didn't merely augment Europe; it reshaped it. Similarly, digital technologies don't merely enhance our traditional literacies; they redefine them.

The essence of this transformation is evident in the arena of digital literacy. The internet's emergence was not simply an alternative medium for traditional content delivery. Instead, for many, it heralded a paradigm shift in the art of reading, writing, and research. The challenge and opportunity were to comprehend and adapt to these new digital literacies, which include online research, content creation across multiple media, and writing for online communication to potentially large audiences.

The rise of ChatGPT reinvigorates debates around the nature of writing and the impact of technology on authorship. Does it signify a mere shift in how writing is done, or does it denote a fundamental metamorphosis of the nature of writing? Our exploration of Kailing's engagement with ChatGPT offers valuable insights into how one adept user of technology navigated this liminal space, authoring through interaction with generative AI.

Our paper highlights questions regarding the nature of writing proficiency and sheds light on the potential of tools such as ChatGPT in academic settings. The case of Kailing serves as an illustrative microcosm of this broader landscape. As demonstrated, while Kailing utilized ChatGPT as a generative tool, she never abdicated her agency, ensuring that the content produced echoed her academic integrity and authorial voice.

This intricate balance between humans and tools, mediated by contextual rules, communities, and divisions of labor (as framed by activity theory), forms the crux of our study. While the "blind man and the stick" analogy underscores the interdependency of humans and tools, Kailing's narrative reveals a nuanced engagement, where ChatGPT serves both as a collaborator and a tool, aiding but never replacing the human intellect.

To this end, our study highlights Kailing's changing relationship with ChatGPT in academic writing, focusing on the tool's strengths and limitations. While Kailing appreciated ChatGPT's assistance in employing sentence variation, she noted recurring patterns that limited its utility. Her journey underscores the balance between generative AI support and the indispensable value of human expertise in refining and enhancing writing.

The crux of our exploration is not just how AI tools such as ChatGPT can be employed, but more critically, how they reshape the writing process, challenge traditional notions of authorship, and make us rethink the nature of writing pedagogy and assessment in academic settings. At the same time, Kailing's continued journey underscores the relationship between AI and the human experience, underscoring that while tools such as ChatGPT can offer invaluable support, their limitations necessitate human intervention to ensure high-quality writing.

**Conclusion**

We present a case study of one participant, understanding its limitations. Kailing's processes and results may well stem largely from her own intellect, personality, or other idiosyncratic characteristics. We do not argue that these results are universal.

However, especially at the emergence of new phenomena, small case studies can help us develop hypotheses to explore through replications and larger studies. Kailing's example suggests new possibilities for human-AI collaboration that can be of great benefit for second language writing and writers, despite generative AI's biases, inaccuracies, and limitations.

Further work is warranted to understand how applicable this case is to the broader field of second language writing with technology, and what kinds of pedagogy are needed to help other students learn to critically exploit this powerful new tool.

**Acknowledgments**

We would like to thank the National Science Foundation (Grant 23152984) for providing the funding that made this project possible. The findings expressed in this work are those of the authors and do not necessarily reflect the views of the National Science Foundation.

**Declaration of Interest**

The authors have no competing interests to declare.